\documentclass[11pt]{article}
\PassOptionsToPackage{table,dvipsnames}{xcolor}
\usepackage[final]{acl}

\usepackage{times}
\usepackage{float}
\usepackage{latexsym}
\usepackage{enumitem}
\usepackage{tabularx}
\usepackage{lmodern}
\usepackage{anyfontsize}
\usepackage[T1]{fontenc}
\usepackage[utf8]{inputenc}
\usepackage{microtype}
\usepackage{inconsolata}
\usepackage{graphicx}
\usepackage{newtxtext,newtxmath}
\usepackage{booktabs,multirow,longtable,array}
\usepackage{wrapfig}
\usepackage{fvextra}
\usepackage[skins,breakable]{tcolorbox}
\usepackage{adjustbox}
\usepackage{pifont}
\usepackage{subcaption}
\usepackage{placeins}

\DeclareUnicodeCharacter{2011}{\mbox{-}}
\DeclareUnicodeCharacter{2192}{\ensuremath{\rightarrow}}
\fvset{
  frame=none, framesep=6pt, bgcolor=gray!3,
  fontsize=\footnotesize, breaklines, breakanywhere,
  breaksymbolleft={}, breaksymbolright={}, numbers=none
}
\tcbset{
  colback=gray!2, colframe=gray!35, boxrule=0.4pt, arc=2mm,
  left=6pt, right=6pt, top=6pt, bottom=6pt
}
\newtcolorbox{promptbox}[1][]{
  enhanced, breakable,
  colback=blue!2, colframe=blue!40!black,
  boxrule=0.5pt, arc=3pt,
  left=8pt, right=8pt, top=6pt, bottom=6pt,
  fonttitle=\bfseries\small, title={#1}, before upper={\small}
}
\newcommand{\cmark}{\textcolor{green!70!black}{\ding{51}}}

\title{\raisebox{-.5cm}{\includegraphics[width=1.5cm]{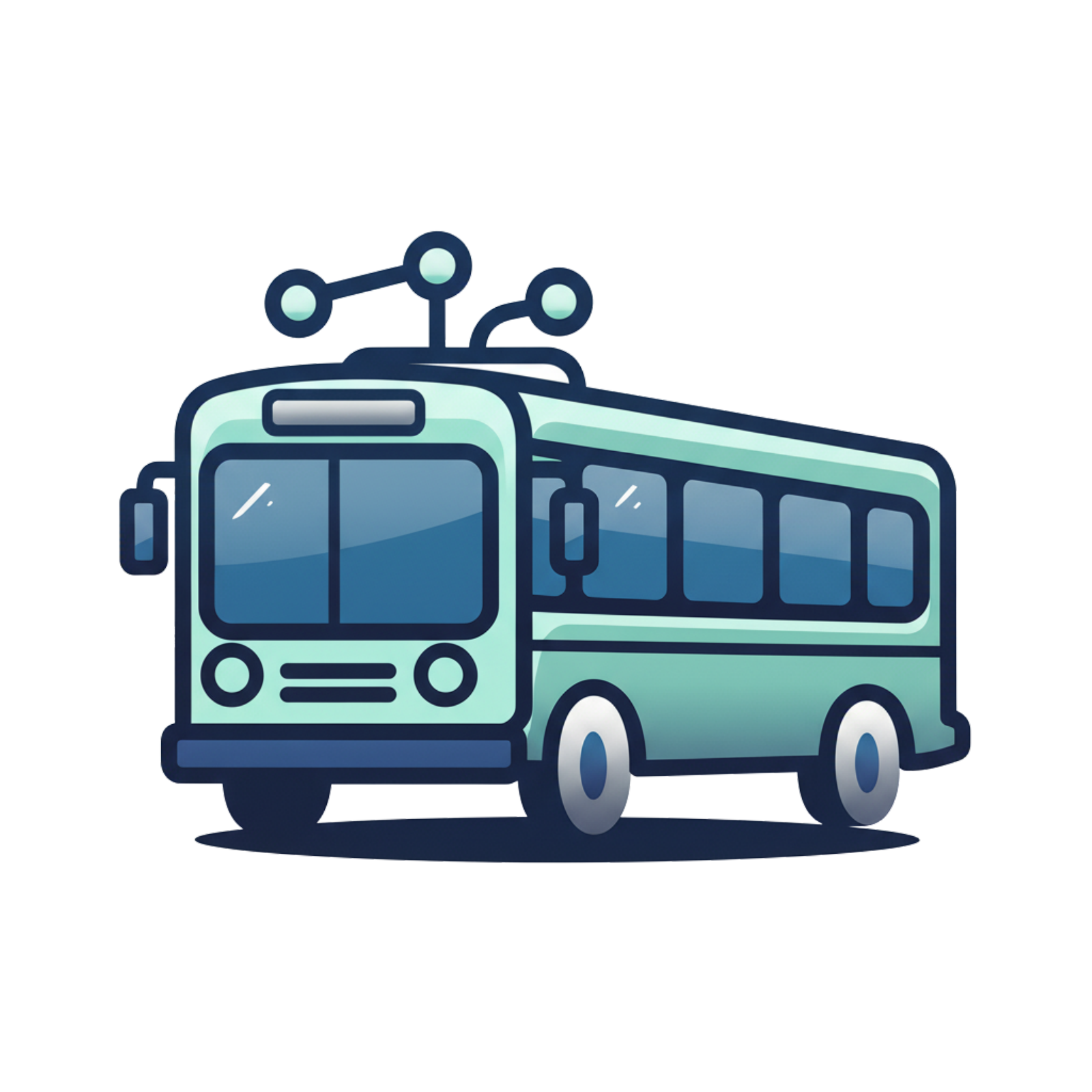}}
\textsc{BusMA}: A Bus Communication Substrate for Multi-Agent Systems}

\author{Yanwen Peng \quad Delvin Ce Zhang \quad Xi Wang \quad Nikolaos Aletras
\\
Department of Computer Science, The University of Sheffield \\
Sheffield, United Kingdom \\
\texttt{\{ypeng86, delvin.ce.zhang, xi.wang, n.aletras\}@sheffield.ac.uk}}

\begin{document}
  \maketitle

\begin{abstract}
Multi-Agent (MA) systems are effective at solving complex tasks that demand planning, tool use, and the synthesis of evidence from multiple sources. Existing systems typically adopt Hierarchical Manager-Worker (HMW) or Router-based Message Passing (RMP) structures as their communication protocol. However, these designs restrict agent autonomy: Worker agents cannot directly consult specific ``peers'', and misrouted messages can propagate errors. Inspired by bus architectures in computer systems, we propose \textsc{BusMA}, a communication framework that allows any agent to address other agents through a shared channel, i.e., the \textit{Bus}. It consists of agent registration, message routing, and shared memory management components.
\textit{Worker agents}, each equipped with tools, have their own local memory and can reason, act (tool usage), and communicate by posting shared messages with specific intents. We introduce four intents: discussion, challenge, guidance, and request for explanation, which support fine-grained communication among agents. A \textit{Chair agent} monitors the shared memory to coordinate interactions and facilitate convergence among Workers. To evaluate the effectiveness of BusMA, we conduct extensive experiments with two frontier LLMs across 13 tasks spanning visual reasoning, mathematical reasoning, and knowledge retrieval. The results demonstrate that \textsc{BusMA} consistently outperforms state-of-the-art HMW and RMP methods.\footnote{Code: \url{https://github.com/YanwenPneg/BusMA}.}
\end{abstract}

\section{Introduction}
\label{sec:introduction}
Multi-Agent (MA) systems powered by Large Language Models (LLMs) \citep{openai_gpt52_2025,deepmind_gemini3pro_nd, deepseekai2025deepseekv3technicalreport}, comprise a set of agents that can reason, act, and communicate to solve complex real-world tasks. These tasks, such as mathematical reasoning \citep{lei2024macm} and knowledge retrieval \citep{huang2025deepresearchagentssystematic}, require effective collaboration \citep{fang2025comprehensivesurveyselfevolvingai,du2025surveyoptimizationlargelanguage}. Hence, the effectiveness of MA systems largely depends on the communication quality between agents and their coordination  \citep{10.24963/ijcai.2024/890,chen2023agentverse,liang-etal-2024-encouraging,du2024improving,wu2023empirical,peng2026statebridge}.

\begin{figure}[t]
    \centering
    \includegraphics[width=1\linewidth]{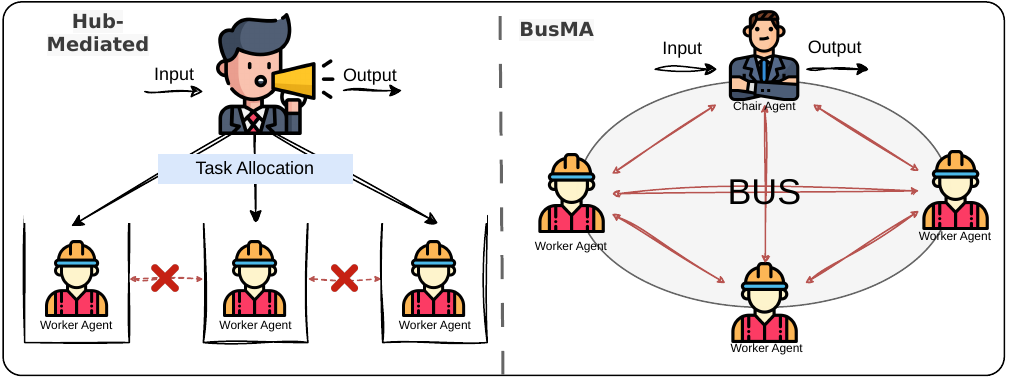}
    \caption{Comparison of hub-mediated topologies and BusMA. 
\textbf{Left}: In hub-mediated frameworks, Workers have restricted autonomy as passive executors, unable to initiate worker-to-worker communication (indicated by \texttimes). 
\textbf{Right}: BusMA enables agents to directly address specific peers through a shared channel (the Bus), supporting four communication intents: Discussion, Challenge, Guidance, and Request for Explanation.}
    \label{fig:framework_compare}
\end{figure}

MA frameworks typically rely on \textit{hub-mediated topologies}, which use a central component to route communication between worker agents, such as Hierarchical Manager-Worker (HMW) and Router-based Message Passing (RMP). HMW employs a \textit{manager agent} to assign tasks and collect results~\citep{qian-etal-2024-chatdev,gu2025agentgroupchat,zhang2025agentorchestrahierarchicalmultiagentframework}. Similarly, RMP uses a \textit{router agent} to mediate all message exchanges~\citep{wu2024autogen,nonomura2025speaksnextmultipartyai}. However, these topologies have two main limitations: (1) RMP and HMW agents have \textbf{restricted autonomy} as passive executors, unable to initiate worker-to-worker communication for critique or error correction~\citep{ke2025surveyfrontiersllmreasoning,jin2025comprehensive,krishnan2025advancing}; and (2) the HMW manager and RMP router suffer from \textbf{error propagation}, acting as \textit{cognitive bottlenecks} that force all information to pass through a \textit{single} hub~\citep{piatti2024cooperate,han2024llm,maragheh2025future}.

To overcome these limitations, we draw inspiration from distributed systems~\citep{oki1993information} and the computer bus architecture~\citep{patterson2017computer} to propose \textsc{BusMA}, a communication substrate that enhances agent autonomy, allowing collaborative interactions on demand. Analogous to a hardware bus where components independently initiate data transfers, \textsc{BusMA} enables agents to directly address specific peers through a shared channel (i.e., the Bus), rather than passively waiting for task assignments or router mediation. BusMA also follows a logic similar to the Information Bus proposed by~\citet{oki1993information} where distributed systems share a common communication channel by maintaining autonomy. Figure \ref{fig:framework_compare} illustrates existing frameworks and BusMA for MA communication. 

The Bus includes three modules: \textit{agent registration} to register agent roles and unique addresses, \textit{message routing} for facilitating direct communication, and \textit{shared memory} (history) management. BusMA instantiates a set of Worker agents and a Chair agent. Each Worker is equipped with tools, maintains its own local memory and can reason, act (tool usage), and communicate through the Bus. Unlike current HMW and RMP approaches, when a Worker needs to communicate with a specific peer, it simply posts a message to the Bus. Messages cover four communication intents: discussion, challenge, guidance, and request for explanation. This keeps agent-specific reasoning local, while only the necessary information is shared through the Bus. The Chair agent, a special case of a Worker agent with act disabled, monitors the shared memory to identify conflicts, detect missing evidence, and issue  follow-up requests, to facilitate convergence. Compared to HMW approaches, the Chair agent acts as \textit{primus inter pares} (first among equals), mitigating information bottlenecks. 

Extensive experiments across 13 datasets show that BusMA outperforms competitive HMW and RMP frameworks. Our analysis reveals that collaboration yields larger gains on tasks where intermediate steps can be verified by peers, such as mathematical reasoning and knowledge retrieval, while producing competitive performance on perceptual tasks such as visual reasoning. Our analysis further shows that fine-grained role division is not 
uniformly beneficial and that Worker heterogeneity can allow errors to propagate undetected.


\section{Related Work}
\label{sec:related_work}

\subsection{LLM-based Agents.}

LLMs such as GPT \citep{achiam2023gpt}, Gemini \citep{comanici2025gemini25pushingfrontier}, and DeepSeek \citep{deepseekai2025deepseekv3technicalreport} serve as the backbone for autonomous agent development. Agentic architectures augment the base LLM with complementary mechanisms, including advanced planning strategies for task decomposition \citep{huang2024understandingplanningllmagents,li2025agentoriented,hu2025agentgen,erdogan2025planandact, zhou2024isr,zhang2025aflow}, external tool use and knowledge bases \citep{zhang2024multimodal,wu2024avatar,qin2024toolllm,feng2025retoolreinforcementlearningstrategic}, and long-term memory or reflection mechanisms for persistent state and iterative improvement \citep{shinn2023reflexion,zhong2024memorybank,mei2024aios,xu2025amem,chhikara2025mem0buildingproductionreadyai}. LLM-based agents have been applied to diverse domains such as automated programming \citep{trivedi-etal-2024-appworld,zhang-etal-2024-codeagent,chen-etal-2025-locagent}, system interaction \citep{wu2024oscopilot,bonatti2025windows}, and scientific discovery \citep{hong2024data,novikov2025alphaevolve}. However, single-agent systems remain brittle on long-horizon, interdependent tasks, with evaluations reporting systematic failures in planning, decision-making, and instruction following \citep{liu2024agentbench,10.5555/3692070.3694316,wang2025odysseybenchevaluatingllmagents,mohammadi2025evaluation,chen2025mlrbench}. 
However, effective dialogue and collaboration between specialized agents is important for improving downstream performance \citep{li2023camel,du2024improving,tran2501multi,chen2412survey}.

\subsection{LLM-based MA Systems.} 
HMW and RMP topologies rely on a hub (a manager or a router) for coordination and information flow. Representative HMW systems include AgentVerse and ChatDev~\citep{chen2023agentverse,qian-etal-2024-chatdev,qian2025scaling,10.1145/3712003}, tool-oriented approaches such as OctoTools~\citep{lu2025octotools}, and evolving orchestration paradigms that train a centralized orchestrator via reinforcement learning~\citep{dang2025chatdev2}. RMP systems include AutoGen~\citep{wu2024autogen} and router-based variants~\citep{yue-etal-2025-masrouter,yao2025toward,nonomura2025speaksnextmultipartyai}, with recent work introducing learnable protocol routing for scenario-aware selection~\citep{du2025protocolrouter}. However, communication between agents in RMP and HMW is restricted~\citep{ke2025surveyfrontiersllmreasoning,jin2025comprehensive,krishnan2025advancing}; while these frameworks may also suffer from error propagation \citep{piatti2024cooperate,han2024llm,maragheh2025future,sagirova2024shared}. These limitations highlight the need for more flexible communication strategies.

\begin{figure*}[ht]
    \centering
    \includegraphics[width=0.8\linewidth]{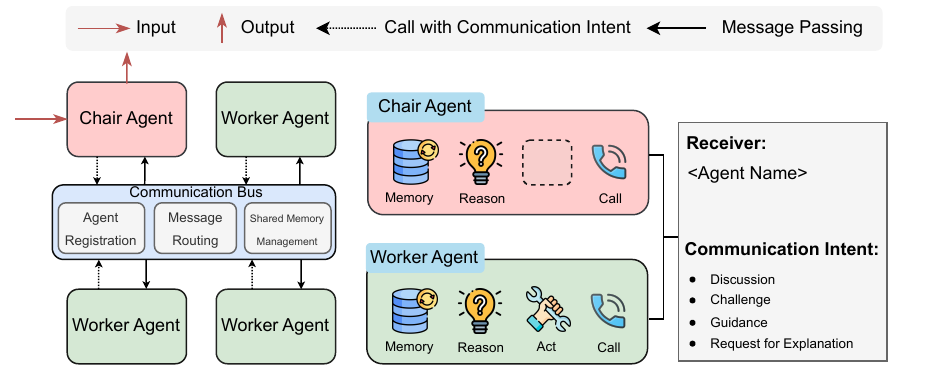}
    \caption{Overview of the BusMA communication substrate.}
    \label{fig:Architecture}
\end{figure*}

\section{The BusMA Framework}

Our \textsc{BusMA} framework consists of three components: (1) \textbf{Worker Agents}, agents equipped with tools that maintain their own local memory and can reason, act (tool usage), and communicate; (2) a \textbf{Chair Agent}, a special case of a Worker agent with action execution disabled; and (3) the \textbf{Bus}, the infrastructure handling agent registration, message routing, and shared memory management. Figure~\ref{fig:Architecture} illustrates this architecture.

\paragraph{Workflow Overview.}
BusMA operates through on demand collaboration (Figure~\ref{fig:Architecture}). A task begins with the input to the Chair Agent, who reasons over the task to identify the first Worker to contact and monitors the shared memory. Activated Workers execute an iterative loop: reasoning over local memory, followed by either invoking a local tool (\texttt{Act}) or posting a message to the Bus (\texttt{Call}). A \texttt{Call} generates a message and shares it via the Bus to the target agent; the calling Worker then suspends until reactivated. The Chair is reactivated only when a message is addressed to it. Finally, the Chair accesses the shared memory to assess progress, issue follow-up calls, or complete the task.

\subsection{Worker Agent}
\label{ssec:worker}
Workers are the core operational units of BusMA. Each one is equipped with a specific tool set (e.g., web search, code execution) and communicates by posting shared messages with specific intents via the bus. The main mathematical notations used by BusMA are listed in Table~\ref{tab:agent_notation}.

\begin{table}[t]
    \centering
    \scriptsize
    \begin{tabularx}{\columnwidth}{@{}c@{\hspace{10em}}X@{}}
        \toprule 
        \textbf{Symbol} & \textbf{Description} \\
        \midrule 
        \multicolumn{2}{@{}l}{\textit{Worker Agent $\alpha_{i}$}} \\
        $\mathcal{L}$ & LLM backbone \\
        $\mathcal{P}_{i}$ & System prompt \\
        $\mathcal{M}_{i}$ & Messages from the bus \\
        $\mathcal{K}_{i}(t)$ & Local memory at iteration $t$ \\
        $\mathcal{T}_{i}$ & Available tool set \\
        $\mathcal{N}$ & Accessible peer agents \\
        $T_{\max}$ & Max local iterations per activation \\
        $r_{i,t}^{\mathrm{priv}}$ & Private reasoning at iteration $t$ \\
        \midrule
        \multicolumn{2}{@{}l}{\textit{Chair Agent ($\mathcal{T}_{\text{chair}} = \emptyset$)}} \\[3pt]
        $\mathcal{P}_{\text{chair}}^{\mathrm{COOR}}$ & Coordination mode prompt \\[3pt]
        $\mathcal{P}_{\text{chair}}^{\mathrm{SUBM}}$ & Submission mode prompt \\[3pt]
        $r_{\text{chair},t}^{\mathrm{glob}}$ & Global view at iteration $t$ \\[3pt]
        \midrule
        \multicolumn{2}{@{}l}{\textit{Communication Bus}} \\
        $s$ & Routing index  \\
        $\mathcal{A}_{i}$ & Unique agent address \\
        $\mathcal{R}$ & Address registry \\
        $\mathcal{Q}_{i}$ & Message queue (FIFO buffer) \\
        $\mathcal{H}$ & Shared memory (history) \\
        \bottomrule
    \end{tabularx}
    \caption{Mathematical notation for BusMA components.}
    \label{tab:agent_notation}
\end{table}
\paragraph{Activation and Initialization.} When a Worker receives a message from the Bus, it is activated and begins its iteration ($t=0$). It constructs its initial context by concatenating the system prompt $\mathcal{P}_{i}$, the received message $\mathcal{M}_{i}$, and descriptions of peers $\mathcal{N}$ and tools $\mathcal{T}_{i}$. By modifying $\mathcal{P}_{i}$ and $\mathcal{T}_{i}$, BusMA instantiates workers with different capabilities tailored to specific tasks (e.g., search, code execution). During execution, the Worker maintains a local memory state $\mathcal{K}_{i}(t)$ that accumulates private reasoning and tool output across subsequent iterations.

\paragraph{Reasoning and Action Selection.} Following the ReAct framework \citep{yao2023react}, each Worker alternates between reasoning and acting. At iteration $t$, the LLM backbone $\mathcal{L}$ analyzes the current context (including $\mathcal{M}_{i}$ and $\mathcal{K}_{i}(t)$) to produce private reasoning $r_{i,t}^{\mathrm{priv}}$, capturing how it interprets new information and selects the next step. Based on $r_{i,t}^{\mathrm{priv}}$ and the available resources, the Worker selects one of two actions: \texttt{Act} or \texttt{Call}. \texttt{Act} invokes a tool to gather information or perform operations; \texttt{Call} posts a shared message to the Bus.

\begin{itemize}
\item \textbf{Act.} The Worker identifies a tool $\tau_{j}\in \mathcal{T}_{i}$ and configures its execution arguments $\theta$ (e.g., a file path or a formulated search query). The Worker executes the tool and receives an output $o_{t}= \tau_{j}(\theta)$. For instance, a search tool may take a query as $\theta$ and return a list of publications as the output. The Worker then updates its local memory with both the private reasoning and the tool output,
\begin{equation}
\mathcal{K}_{i}(t+1) = \mathcal{K}_{i}(t) \cup \{ r_{i,t}^{\mathrm{priv}}, o_{t}\}. \label{eq:worker_memory_update}
\end{equation}
Next, it proceeds to iteration $t+1$, reconstructing its context with $\mathcal{P}_{i}$, the latest Bus message $\mathcal{M}_{i}$, the peer and tool descriptions $(\mathcal{N}, \mathcal{T}_{i})$, and the updated local memory $\mathcal{K}_{i}(t+1)$.

\item \textbf{Call.} The Worker chooses a target agent $\alpha_{j}\in \mathcal{N}$ and generates an action $a_{t}^{\mathrm{call}}= (\alpha_{j}, m_{i \to j})$, where $m_{i \to j}$ is the shared message. The Worker formats the message as JSON, and the Bus parses it for sending it to the target agent. After posting the message to the Bus, the Worker suspends its current iteration.
\end{itemize}

To enable high-quality communication, each Worker frames $m_{i\to j}$ by selecting a communication intent. We define four types: \textbf{discussion} for bidirectional information exchange, \textbf{request for explanation} for clarifying ambiguous context, \textbf{challenge} for questioning results, and \textbf{guidance} for providing expertise. The Worker pairs the intent with corresponding evidence and a focused question (e.g., tool output or a detected inconsistency). These four intent types expand the communication space beyond conventional assign-and-return exchanges in hub-mediated systems. In particular, the challenge intent allows one agent to question another's intermediate results, enabling re-examination through deliberation rather than passive error propagation.

The Worker's current iteration terminates under two conditions: (1) the Worker chooses \texttt{Call}, or (2) the Worker reaches the maximum iteration limit $T_{\max}$. In the second case, the Worker posts a failure message to the agent that originally activates it. Appendix~\ref{ssec:prompt_abstract} provides an example prompt for the Worker, including the JSON communication protocol and formatting requirements.

\subsection{Chair Agent}
\label{ssec:chair_agent}
The Chair Agent is a special case of a Worker agent with \texttt{Act} disabled, serving as the system's coordinator and entry point. Unlike existing HMW and RMP systems, the Chair Agent does not generate an explicit plan or decompose and assign subtasks upfront. Instead, it initiates collaboration and synthesizes information from Worker interactions. BusMA decouples message routing from task synthesis: Workers address specific peers directly on the Bus, and this peer-to-peer information exchange completely bypasses the Chair.
The Chair does not use tools (i.e., $\mathcal{T}_{\text{chair}}= \emptyset$, see Figure~\ref{fig:Architecture}), instead it operates in two modes with different prompts: $\mathcal{P}_{\text{chair}}^{\mathrm{COOR}}$ for the coordination phase and $\mathcal{P}_{\text{chair}}^{\mathrm{SUBM}}$ for the submission phase. Appendix~\ref{ssec:prompt_chair} presents the two-phase prompts for the Chair agent.

\paragraph{Coordination Mode.} At $t=0$, the Chair receives the task and begins coordination using $\mathcal{P}_{\text{chair}}^{\mathrm{COOR}}$. It reasons over the task description to identify the first Worker to contact, then issues a \texttt{Call} before it starts monitoring the shared memory. During this period, Workers communicate with each other through \texttt{Call}, exchanging information, critiques, and requests.

The Chair is reactivated only when a Worker  addresses it. Upon reactivation, the Chair accesses the shared message (history) $\mathcal{H}$ to update its global view $r_{\text{chair},t}^{\mathrm{glob}}$ of task progress. Specifically, it identifies unresolved disagreements, detects missing evidence or unaddressed questions, and formulates follow-up requests. Then, the Chair either (i) continues coordination by calling another agent, or (ii) outputs a JSON structure with a \texttt{submit} field to transition to the submission mode.

\paragraph{Submission Mode.} The Chair switches to $\mathcal{P}_{\text{chair}}^{\mathrm{SUBM}}$ and generates the final response by synthesizing the task description and accumulated information. This is a single-step generation.

\subsection{Communication Bus}
\label{ssec:bus}
The communication Bus provides a shared substrate for agent interaction with three modules: \textit{Agent Registration}, \textit{Message Routing}, and \textit{Shared Memory Management}.

\paragraph{Agent Registration.} Upon system  initialization, each instantiated agent $\alpha_{i}$ (Workers and the Chair) is registered to the Bus and assigned a unique address $\mathcal{A}_{i}$. This creates an address registry $\mathcal{R}= \{(\alpha_{i}, \mathcal{A}_{i}) \mid \alpha_{i}\in \mathcal{N}\}$, enabling agents to be addressed as message receivers.

\paragraph{Message Routing.} To support seamless collaboration, the Bus maintains a dedicated First-In-First-Out buffer $\mathcal{Q}_{i}$ for each agent. When an agent $\alpha_{i}$ issues a \texttt{Call} and posts a message specifying $a_{t}^{\mathrm{call}}= (\alpha_{j}, m_{i \to j})$, the Bus parses the message, extracts the receiver $\alpha_{j}$, and verifies it against the registry $\mathcal{R}$. The Bus then augments the message:
\begin{equation}
\hat{m}_{i \to j}(s)= (m_{i \to j}, \text{ID}, s, \alpha_{i}), \label{eq:bus_augmented_message}
\end{equation}
where \text{ID} is a unique identifier, $s$ is the routing index assigned by the Bus, and $\alpha_{i}$ is the sender. The Bus pushes $\hat{m}_{i \to j}$ into the receiver queue $\mathcal{Q}_{j}$. The receiver consumes messages in arrival order. Upon activation, an agent pops one message from its queue $\mathcal{Q}_{i}$, and the remaining messages await subsequent activations. This design allows an agent to post messages to multiple receivers concurrently while maintaining  consistency, as each receiver consumes incoming messages in a strict order.

\paragraph{Shared Memory Management.} The Bus maintains a shared memory (history) $\mathcal{H}$ that stores all shared messages routed. Each newly sent message is appended to $\mathcal{H}$ in ascending order of its routing index $s$ (with the unique ID assigned at routing time), preserving the complete communication history. When any Worker posts a message to the Chair, the Bus provides the Chair with access to $\mathcal{H}$, enabling global progress assessment and synthesis during coordination.

\begin{figure}[t]
\centering
\includegraphics[width=0.9\linewidth]{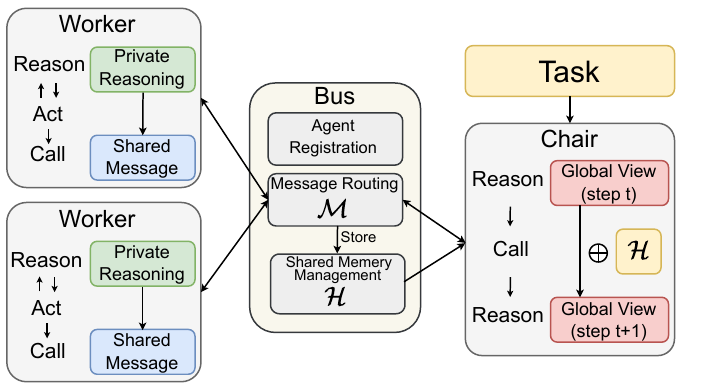}
\caption{Information flow in BusMA: private reasoning $r_{i,t}^{\mathrm{priv}}$, shared message $m_{i \to j}$, and global view $r_{\text{chair},t}^{\mathrm{glob}}$.}
\label{fig:insight_formation}
\end{figure}

\subsection{Overall Information Flow in BusMA}
\label{ssec:insight}
We propose a new method to organize information at three levels of granularity (Figure \ref{fig:insight_formation}). Each Worker maintains private reasoning $r_{i,t}^{\mathrm{priv}}$ in its local memory $\mathcal{K}_{i}(t)$, accumulating intermediate evidence and tentative conclusions that remain internal unless made public to the Bus via \texttt{Call}. When a Worker issues a \texttt{Call}, it publicizes only actionable content as a shared message $m_{i \to j}$ with a communication intent (\S\ref{ssec:worker}) and supporting evidence. Upon each reactivation, the Chair accesses $\mathcal{H}$ to update its global view $r_{\text{chair},t}^{\mathrm{glob}}$, identifying conflicts, detecting gaps, and issuing targeted follow-ups. We hypothesize that this separation reduces communication overhead while preserving coordination coherence.

\section{Experimental Setup}

\subsection{Benchmarks}
Following \citet{lu2025octotools} and \citet{smolagents}, we evaluate \textsc{BusMA} on two complementary benchmark groups: one prioritizing breadth across modalities and the other emphasizing depth in complex problem-solving.

\paragraph{Diversity-oriented.} We use 12 standard datasets spanning three categories: (1) \textit{Visual Reasoning} (e.g., VQA 2.0), (2) \textit{Mathematical Reasoning} (e.g., Omni-MATH), and (3) \textit{Knowledge Retrieval} (e.g., HotpotQA). Table \ref{datasets} details the domain, modality, and required skills for each task. Following \citet{lu2025octotools}, we adopt a sample size of 200 instances per dataset and apply their corresponding evaluation metrics for a fair comparison. Detailed dataset description is provided in Appendix \ref{sec:experiments_setup}.

\begin{table}[t]
\centering
\Large
\resizebox{0.95\columnwidth}{!}{%
\setlength{\tabcolsep}{3pt}
\renewcommand{\arraystretch}{1.1}
\begin{tabular}{@{}l l l c c c c@{}}
\toprule
\textbf{Datasets} & \textbf{Modality} & \textbf{Domain} & \raisebox{-0.2\height}{\includegraphics[height=8pt]{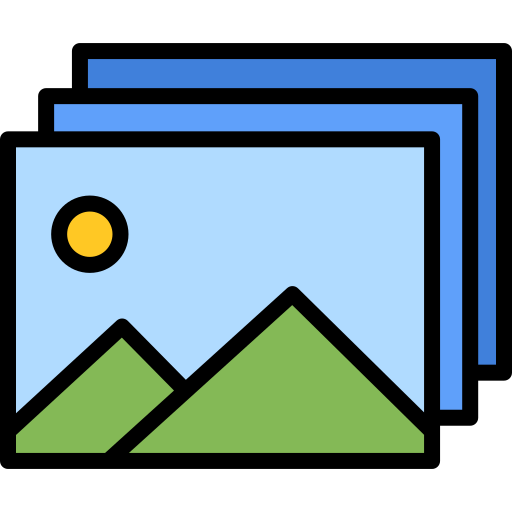}} & \raisebox{-0.2\height}{\includegraphics[height=8pt]{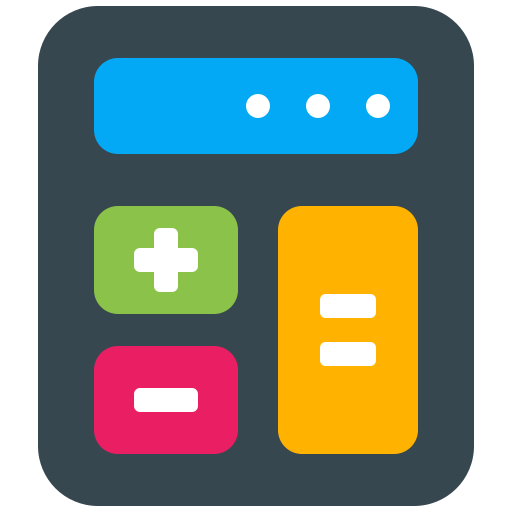}} & \raisebox{-0.2\height}{\includegraphics[height=8pt]{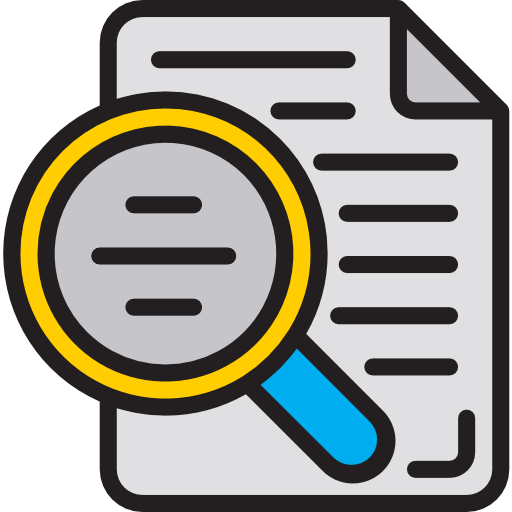}} & \raisebox{-0.2\height}{\includegraphics[height=8pt]{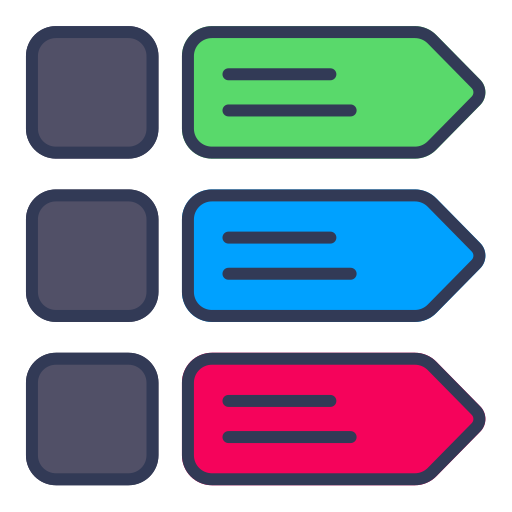}} \\
\midrule
AlgoPuzzleVQA \citep{ghosal-etal-2025-algopuzzlevqa} & Vision & General & \cmark & & & \cmark \\
Hallusion-VD \citep{guan2024hallusionbench} & Vision & General & \cmark & & & \\
PuzzleVQA \citep{chia2024puzzlevqa} & Vision & General & \cmark & & & \cmark \\
VQA 2.0 \citep{goyal2017making} & Vision & General & \cmark & & & \cmark \\
\midrule
Game of 24 \citep{nlile_24_game_2025} & Text & Math & & \cmark & & \cmark \\
Omni-MATH \citep{gao2024omni} & Text & Math & & \cmark & \cmark & \cmark \\
CLEVR-Math \citep{lindstrom2022clevr} & Vision & Math & \cmark & \cmark & & \\
MathVista \citep{lu2023mathvista} & Vision & Math & \cmark & \cmark & \cmark & \cmark \\
\midrule
GPQA \citep{rein2024gpqa} & Text & Knowledge & & \cmark & \cmark & \cmark \\
MMLU-Pro \citep{wang2024mmlu} & Text & Knowledge & & & \cmark & \cmark \\
SciFIBench \citep{roberts2024scifibench} & Vision & Knowledge & \cmark & & \cmark & \\
HotpotQA \citep{yang2018hotpotqa} & Text & Knowledge & & & \cmark & \cmark \\
\bottomrule
\end{tabular}%
}
\caption{Diversity-oriented tasks. Icons denote required skills: visual understanding {\includegraphics[height=7pt]{Icons/image.png}}, numerical calculation {\includegraphics[height=7pt]{Icons/calculator.png}}, knowledge retrieval {\includegraphics[height=7pt]{Icons/search.png}}, and multi-step reasoning {\includegraphics[height=7pt]{Icons/list-option.png}}.}
\label{datasets}
\end{table}

\paragraph{Complexity-oriented.} We use the \textbf{GAIA} \citep{mialon2024gaia} validation set (165 questions). This benchmark requires agents to combine file parsing, web browsing, and code execution. Following the OpenDeepResearch \citep{smolagents} setting, we report exact-match accuracy.

\subsection{Baselines}
For \textbf{diversity-oriented tasks}, we evaluate four general-purpose MA frameworks: \textit{OctoTools} \citep{lu2025octotools} (HMW-style planner–executor), \textit{SmolAgents} \citep{smolagents}, \textit{LangGraph} \citep{langgraph} (HMW), and \textit{AutoGen} \citep{wu2024autogen} (RMP). For the \textbf{complexity-oriented GAIA benchmark}, we compare against specialized state-of-the-art systems including \textit{MagenticOne} \citep{fourney2024magenticonegeneralistmultiagentsolving} and \textit{OpenDeepResearch} \citep{smolagents}. To further isolate the benefits of MA collaboration, we also include a single-agent baseline using Gemini-2.5-Flash and Pro with standard Function Calling. Detailed configurations for all baselines are provided in Appendix \ref{sec:baseline_details}.

\begin{table}[!t]
\centering
\Large
\resizebox{1.0\columnwidth}{!}{%
\begin{tabular}{clcccccc}
\toprule
\multicolumn{2}{c}{} & OctoTools & SmolAgents & LangGraph & AutoGen & BusMA (Ours) & $\Delta$ \\
\midrule
\multirow{13}{*}{\rotatebox{90}{DeepSeek\mbox{-}V3}} & AlgoPuzzleVQA & \underline{47.5} & 31.5 & 45.5 & 39.5 & \textbf{60.0} & +12.5 \\
& Hallusion-VD & \underline{73.0} & \textbf{75.5} & 69.0 & 69.5 & 72.5 & -3.0 \\
& PuzzleVQA & \underline{56.5} & 53.0 & 55.5 & 55.5 & \textbf{63.0} & +6.5 \\
& VQA 2.0 & 67.5 & \underline{73.0} & 65.0 & 66.5 & \textbf{75.5} & +2.5 \\
\cmidrule(lr){2-8}
& Game of 24 & \underline{75.0} & 68.5 & 62.0 & 47.5 & \textbf{88.5} & +13.5 \\
& Omni-MATH & \underline{52.0} & 49.5 & 41.0 & 41.0 & \textbf{55.0} & +3.0 \\
& CLEVR-Math & \underline{74.5} & 72.0 & 71.0 & 30.5 & \textbf{77.5} & +3.0 \\
& MathVista & \textbf{65.0} & \underline{63.0} & 53.5 & 55.5 & 62.5 & -2.5 \\
\cmidrule(lr){2-8}
& GPQA & \textbf{60.5} & 54.0 & 55.5 & 45.5 & \underline{56.0} & -4.5 \\
& MMLU-Pro & 68.0 & \underline{73.0} & 52.5 & 59.0 & \textbf{79.5} & +6.5 \\
& SciFIBench & 72.0 & 66.0 & \underline{75.0} & 70.0 & \textbf{76.5} & +1.5 \\
& HotpotQA & 53.5 & \underline{54.5} & 30.5 & 50.5 & \textbf{57.0} & +2.5 \\
\rowcolor{gray!10}
& \textbf{Average} & \underline{63.8} & 61.1 & 56.3 & 52.5 & \textbf{68.6} & +4.8 \\
\midrule
\multirow{13}{*}{\rotatebox{90}{Gemini-2.5-Flash}} & AlgoPuzzleVQA & \textbf{66.0} & 55.5 & 52.0 & 37.0 & \underline{63.0} & -3.0 \\
& Hallusion-VD & \underline{75.5} & 75.0 & 74.0 & 72.0 & \textbf{77.0} & +1.5 \\
& PuzzleVQA & \textbf{80.5} & 72.0 & 75.0 & 58.0 & \underline{76.0} & -4.5 \\
& VQA 2.0 & \textbf{77.5} & 71.5 & 75.5 & 69.5 & \underline{76.0} & -1.5 \\
\cmidrule(lr){2-8}
& Game of 24 & 88.0 & \underline{89.5} & 81.5 & 73.5 & \textbf{96.5} & +7.0 \\
& Omni-MATH & 53.5 & \underline{67.0} & 50.0 & 66.0 & \textbf{68.0} & +1.0 \\
& CLEVR-Math & \underline{89.0} & 71.5 & 76.0 & 57.0 & \textbf{89.5} & +0.5 \\
& MathVista & \underline{77.5} & 67.5 & 64.5 & 55.0 & \textbf{79.0} & +1.5 \\
\cmidrule(lr){2-8}
& GPQA & \underline{68.5} & 64.5 & 59.5 & 64.5 & \textbf{69.5} & +1.0 \\
& MMLU-Pro & 72.0 & \underline{77.0} & 62.5 & 46.0 & \textbf{79.0} & +2.0 \\
& SciFIBench & \underline{81.0} & 75.5 & 78.5 & 54.5 & \textbf{82.5} & +1.5 \\
& HotpotQA & \underline{57.0} & 54.5 & 55.0 & 44.5 & \textbf{59.0} & +2.0 \\
\rowcolor{gray!10}
& \textbf{Average} & \underline{73.8} & 70.1 & 67.0 & 58.1 & \textbf{76.3} & +2.5 \\
\bottomrule
\end{tabular}%
}
\caption{Accuracy (\%) of MA frameworks across tasks and models. Bold denotes best method in each task, and the second best is underlined. $\Delta$ shows the performance difference between BusMA and the best baseline.}
\label{tab:general_main}
\end{table}

\subsection{Implementation Details}
We use \textbf{DeepSeek-V3} \citep{deepseekai2025deepseekv3technicalreport} and \textbf{Gemini-2.5-Flash} \citep{comanici2025gemini25pushingfrontier} as LLM backbone $\mathcal{L}$. Since DeepSeek-V3 does not support image input, all frameworks evaluated under the DeepSeek-V3 setting route visual tool calls to Gemini-2.0-Flash; this mapping is applied uniformly across BusMA and all baselines.
Following the principle of role based specialization \citep{qian-etal-2024-chatdev, zhang2025agentorchestrahierarchicalmultiagentframework}, we instantiate task specific Workers: a \textbf{Web Agent} for retrieval, an \textbf{ImageQA Agent} for vision, and a \textbf{Code Agent} for coding. For GAIA, to isolate the impact of Worker performance, we adopt the \textbf{Browser Agent} and file analysis tools from \citet{smolagents} via adapters that map external agent I/O to the Bus protocol. To ensure fair comparison with baselines that execute agents sequentially, BusMA is configured to activate only one agent at a time during experiments. This setting isolates the effect of communication structure from parallelism. We also compare with a single-agent baseline (Appendix~\ref{sec:single_agent}).
The prompts for each agent are listed in Appendix \ref{sec:system_prompt}. Hyperparameters are provided in Appendix \ref{sec:experiments_setup}.

\section{Results}
\label{sec:results}

\subsection{Diversity Benchmarks} 
Table \ref{tab:general_main} shows results across models and tasks. Overall, BusMA achieves the best average accuracy, surpassing the best baseline (OctoTools) by 4.8\% with DeepSeek-V3 and 2.5\% with Gemini-2.5-Flash. 

In visual reasoning (AlgoPuzzleVQA, Hallusion-VD, PuzzleVQA, and VQA 2.0), BusMA ranks first or second in seven of the eight task--backbone combinations. Improvements are larger with DeepSeek-V3 but more modest with the stronger Gemini-2.5-Flash model, suggesting that, on visual reasoning tasks, accuracy is mainly determined by the base model’s image understanding (multimodal) capability, which additional agent communication is unlikely to substantially improve.

\begin{table}[!t]
\centering
\large
\resizebox{\linewidth}{!}{
\begin{tabular}{l c c c c c c}
\toprule
\textbf{Methods} & MA & \textbf{Level 1} & \textbf{Level 2} & \textbf{Level 3} & \textbf{Overall} \\
\midrule
Gemini-2.5-Flash-F/C & & 30.1 & 12.7 & 7.7 & 17.6 \\
Gemini-2.5-Pro-F/C & & 39.6 & 24.1 & 19.2 & 28.2 \\
MagenticOne & \cmark & 52.8 & 36.0 & 15.4 & 38.2 \\
OpenDeepResearch & \cmark & 58.5 & 43.0 & 19.2 & 44.2 \\
BusMA & \cmark & \textbf{60.3} & \textbf{47.6} & \textbf{26.9} & \textbf{48.5} \\
\bottomrule
\end{tabular}%
}
\caption{Performance (\%) on GAIA. F/C denotes single-call function calling (non-agentic); MA systems use Gemini-2.5-Flash. Best is \textbf{boldfaced}.}

\label{tab:deep-research}
\end{table}

In mathematical reasoning, BusMA achieves the best performance in the majority of benchmarks, including Game of 24, Omni-MATH, and CLEVR-Math across both backbone models. On MathVista, it leads with Gemini-2.5-Flash (79.0\%) and remains competitive with DeepSeek-V3. These results suggest that BusMA's diverse communication intents enhance iterative verification capability of agents, potentially mitigating error propagation that commonly occurs in HMW and RMP methods \cite{pan2025why}.

For knowledge-intensive tasks, BusMA performs strongly on MMLU-Pro, SciFIBench, and HotpotQA, achieving the best results with both backbone models. This advantage may stem from BusMA's three-level information organization: Workers keep outputs from web search tools in their local memory $\mathcal{K}_{i}(t)$, externalizing only more important information. This separation mitigates long-context overhead, while the Chair's access to the shared memory enables synthesis across multiple sources. On GPQA, our method achieves the best result with Gemini-2.5-Flash but ranks second with DeepSeek-V3, indicating sensitivity to settings where retrieval provides limited support. In such cases, Workers may make mistakes, which can mislead the Chair's co-ordination.

Gemini-2.5-Flash outperforms DeepSeek-V3, consistent with prior work \citep{comanici2025gemini25pushingfrontier}. However, BusMA yields larger relative improvement with DeepSeek-V3, suggesting better gains when the base model is weaker. 

\subsection{GAIA Benchmark}

Table \ref{tab:deep-research} shows the performance across GAIA difficulty levels. Single-model baselines Gemini-2.5-Flash-FunctionCalling and Gemini-2.5-Pro-FunctionCalling substantially underperform all MA systems, achieving 17.6\% and 28.2\% respectively, suggesting that architectural design rather than model capacity is critical for tackling complex tasks. More specifically, BusMA achieves 60.3\%, 47.6\%, and 26.9\% accuracy for Levels 1, 2, and 3 respectively, with 48.5\% overall. This corresponds to a 4.3\% improvement over OpenDeepResearch. Notably, BusMA achieves a larger lead on Levels 2 and 3 (4.6\% and 7.7\%), which require more advanced reasoning and complex collaboration. This highlights BusMA's ability to provide effective solution paths. Moreover, despite using the exact same Browser Agent as OpenDeepResearch, BusMA still outperforms the latter across all levels, indicating that the improvement is not solely attributed to the Browser Agent, but associated with the Chair's coordination and the Bus modules for Message Routing and Shared Memory Management.

\section{Analysis}
\label{sec:analysis}

\begin{table}[t]
\centering
\scriptsize
\begin{tabular}{l c c c}
\toprule
\textbf{Configuration} & \textbf{PuzzleVQA} & \textbf{MathVista} & \textbf{GPQA} \\
\midrule
BusMA (Full) & \textbf{76.0} & \textbf{79.0} & 69.5 \\
\midrule
w/ Manager & 71.5 {\scriptsize(-4.5)} & 72.0 {\scriptsize(-7.0)} & 68.0 {\scriptsize(-1.5)} \\
\midrule
w/ SmolAgents Workers & 76.0 {\scriptsize(+0.0)} & 77.5 {\scriptsize(-1.5)} & \textbf{70.5} {\scriptsize(+1.0)} \\
SmolAgents Baseline & 72.0 & 67.5 & 64.5 \\
\bottomrule
\end{tabular}
\caption{Ablation study results (\%). \textit{w/ Planner Chair} replaces the Chair agent with the Manager agent, a task decomposition planner. \textit{w/ SmolAgents Workers} substitutes BusMA Workers with those from SmolAgents.}
\label{tab:ablation}
\end{table}

\subsection{Ablation Study} 
\label{ssec:ablation}
We conduct an ablation study to analyze the contributions of BusMA's core components. We select one representative benchmark from each domain: PuzzleVQA (visual reasoning), MathVista (mathematical reasoning), and GPQA (knowledge retrieval). We use Gemini-2.5-Flash for all the experiments. Table \ref{tab:ablation} shows the results.

First, the \textit{w/ HMW Manager} setting converts BusMA into HMW by replacing the Chair with a Manager. Following \citet{wu2024autogen} and \citet{smolagents}, we modify the Chair's coordination mode prompt to function as a centralized coordinator that assigns tasks and collects results (see Appendix~\ref{ssec:prompt_planner}), and disable communication among Workers. This restricts Workers to passive executors unable to initiate peer-to-peer communication for critique or error correction. 
The accuracy drop of 1.5 to 7.0 percentage points across tasks supports the effectiveness of BusMA's communication and coordination design.

Second, we replace BusMA's Worker implementation with the SmolAgents Worker implementation \citep{smolagents} as described in Section~\ref{sec:experiments_setup} (\textit{w/ SmolAgents Workers}), while keeping the Chair and communication substrate unchanged.
We observe that performance remains close to the original BusMA (within 1.5\%), yet this variant exceeds SmolAgents by 4.0\% to 10.0\% across tasks. This gap indicates that the agents communication accounts for the majority of BusMA's improvement, suggesting robustness to different Worker implementations.

\begin{table}[!t]
\centering
\scriptsize
\setlength{\tabcolsep}{3pt}
\begin{tabular}{@{}l p{0.6\columnwidth} r@{}}
\toprule
\multicolumn{2}{@{}l}{\textbf{Step 1: Chair}} & \textit{Discussion} \\
\midrule
\textsc{Reason} & Requires domain knowledge; no tools available. & \\
\textsc{Call} & Delegate to Web agent for symmetry search. & \\
\midrule
\multicolumn{2}{@{}l}{\textbf{Step 2: Web}} & \textit{Discussion} \\
\midrule
\textsc{Reason} & Search symmetry for each candidate. & \\
\textsc{Act} & \texttt{GoogleSearch(``triisopropyl borate symmetry'')} & \\
\textsc{Call} & Triisopropyl borate has C3h. & {\color{red}\ding{55} Error} \\
\midrule
\multicolumn{2}{@{}l}{\textbf{Step 3: Chair}} & \textit{Submit} \\
\midrule
\textsc{Reason} & Cannot verify; select higher confidence option. & {\color{red}\ding{55} Propagated} \\
\textsc{Call} & Submit A. & \\
\bottomrule
\end{tabular}
\caption{Error case from GPQA. Query: \textit{Which molecule has C3h symmetry?} Correct: C. Output: A.}
\label{table:busma_error_case}
\end{table}

\begin{figure*}[t]
\centering
\includegraphics[width=0.815\linewidth]{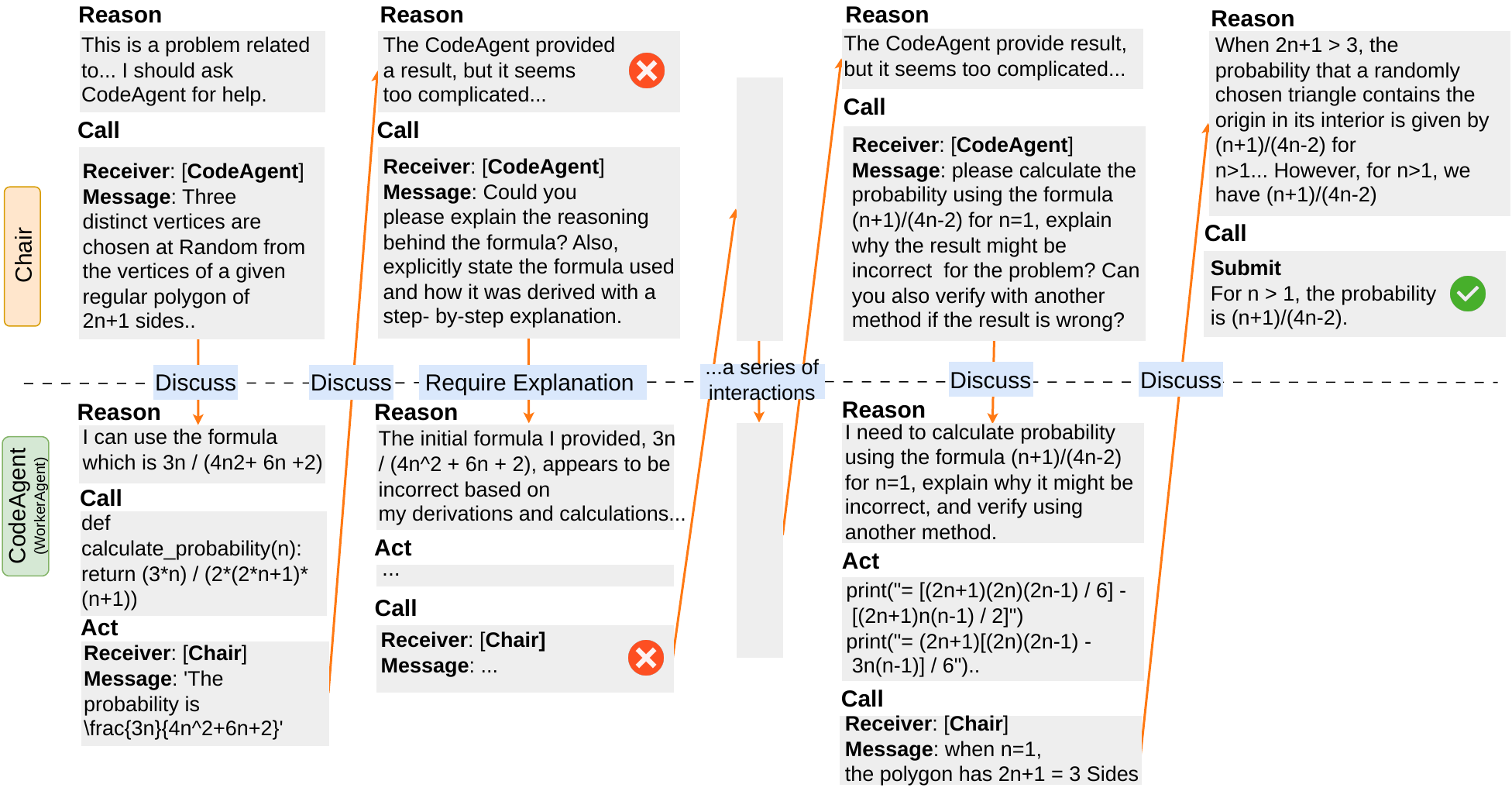}
\caption{BusMA communication trajectory for answering a question from Omni-MATH: \textit{Three distinct vertices are chosen randomly from the vertices of a regular polygon with (2n+1) sides; what is the probability that the center lies inside the triangle formed by the three chosen vertices?} The full trajectory is provided in Table~\ref{table:full_busma_comm_traj}.}
\label{fig:ma_109}
\end{figure*}

\subsection{Qualitative Analysis} 
\label{ssec:qualitative} 
\begin{figure}[t]
\centering
\includegraphics[width=\linewidth]{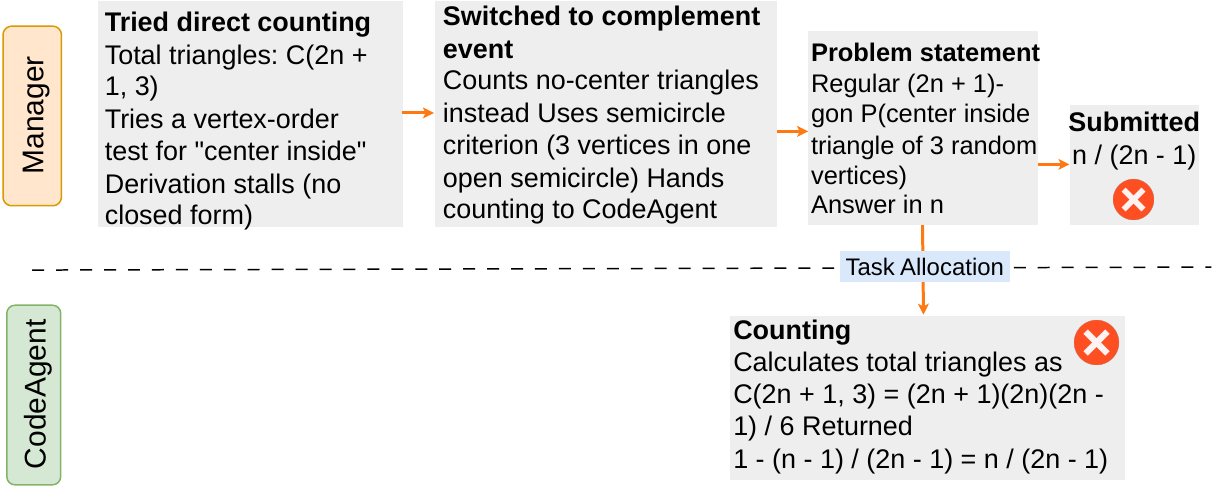}
\caption{SmolAgents communication trajectory for the same Omni-MATH task in Figure \ref{fig:ma_109}.}
\label{fig:ma_109_smol}
\end{figure}

Figures \ref{fig:ma_109} and \ref{fig:ma_109_smol}  show  communication trajectories of BusMA and the best-performing baseline SmolAgents on the same \textbf{Omni-MATH} problem. BusMA tackles this task with two agents: the Chair and the Code Agent. The Chair \textbf{discusses} with the Code Agent about required calculations and \textbf{requests} the Code agent to provide a more detailed explanation after receiving an unreliable response. Looking at the full communication trajectory of BusMA (Table \ref{table:full_busma_comm_traj} in Appendix~\ref{sec:qual_anal_app}), the Chair also \textbf{guides} the Code Agent to validate candidate answers using simple test cases (1, 2 and 3), while the Code Agent \textbf{challenges} the Chair's hypotheses when appropriate. Although both agents initially make errors, iterative interaction yields the correct solution and a brief reflection on the causes of failure. By contrast, for \textbf{SmolAgents}, the Manager generates an incorrect answer and delegates to a Code Agent. However, it only collects feedback without enabling two-way communication, so the initial error persists. Apart from the ablation study in \S\ref{ssec:ablation}, this further highlights the main difference between our Chair Agent acting as \textit{primus inter pares} with Workers compared to standard HMW managers. Appendix~\ref{sec:clevr_analysis} provides an analysis of over-specialization in CLEVR-Math.

\subsection{Error Analysis} 

Table~\ref{table:busma_error_case} presents a failure case from GPQA where BusMA produces an incorrect answer due to erroneous retrieval (a detailed analysis is provided in Appendix~\ref{sec:error_anal_app}). Based on the GoogleSearchTool retrieval outputs, the Web Agent incorrectly claims that \textit{triisopropyl borate has C3h symmetry} (it actually belongs to C3, lacking the horizontal mirror plane), while providing only speculative information for the correct answer. The Chair, lacking access to GoogleSearchTool, cannot independently verify the Web Agent's claims and thus selects the option presented with higher confidence. This case highlights a limitation of Worker heterogeneity: when specialized tools are exclusively assigned to specific agents, other agents have no means to cross-validate their outputs, allowing retrieval errors to propagate undetected.

\section{Conclusion}
We introduced BusMA, a communication substrate that enables agents to directly address specific peers through a shared channel, supporting four communication intents: discussion, challenge, guidance, and request for explanation. BusMA features Worker Agents that can reason, act, and communicate, a Chair Agent that monitors shared memory to facilitate convergence, and a Bus with agent registration, message routing, and shared memory management modules. Extensive experiments across 13 tasks spanning visual reasoning, mathematical reasoning, and knowledge retrieval demonstrate that BusMA consistently outperforms state-of-the-art HMW and RMP methods. In future work, we plan to extend agents to incorporate confidence estimation, allowing them to identify and mitigate information asymmetry. We also aim to evaluate BusMA on tasks such as WebWalkerQA~\citep{wu2025webwalkerbenchmarkingllmsweb} that require coordinating more agents and longer reasoning chains to further assess its scalability.

\clearpage

\section*{Limitations}
BusMA improves collaboration through interaction but does not enhance an LLM's underlying capabilities. In visual reasoning tasks where perception is the bottleneck, additional exchanges may introduce noise, yielding smaller gains or regressions.
A core limitation is error propagation through the shared memory. When external tools provide incomplete or wrong evidence, Workers may produce misleading messages that bias the Chair's synthesis, as seen in our weaker GPQA results. Tool heterogeneity compounds this issue: agents often cannot verify outputs from unfamiliar tools, allowing errors to spread. Confidence-aware arbitration with conflict detection could help address this.

The tasks used in our experiments typically involve coordination among only 2--3 agents. Widely used multi-agent benchmarks do not yet require large-scale agent collaboration, and prior work has found that increasing the number of agents does not necessarily improve task performance. Architecturally, BusMA does not impose structural barriers to scaling: each agent receives a unique address and the shared memory grows linearly with the number of routed messages. As the number of agents increases, the shared memory accumulates more messages and the Chair must synthesize longer communication histories, creating bus contention. Hardware bus mechanisms such as blocking replies, arbitration protocols, and priority scheduling are beyond the scope of this work. In future work, we plan to include these features to enable synchronization in larger-scale scenarios.

\bibliography{custom}

\clearpage

\appendix

\FloatBarrier
\onecolumn
\section{System Prompts}
\label{sec:system_prompt}

Here, we provide the prompts used in our experiments for reproducibility. Each box contains the system prompt for the corresponding agent.

\vspace{0.5em}
\begin{promptbox}[Prompt A: Abstract Prompt]
\label{ssec:prompt_abstract}
\begin{Verbatim}
Role
You are a focused analysis assistant within a multi-agent system. You analyze tasks, use tools, and communicate findings precisely.

Team Structure
You work collaboratively with the following agents:
<Available Agents>

Communicate in different types
<discussion,  Request for explaination, challenge, guidance>

Working Framework
Follow a Reason-act-call loop:
1) Think  2) Act (tool call)  3) Observe  4) Iterate  5) Call 

Output (JSON‑only)
{
  "thought": "<reasoning, strategy, next steps>",
  "action": { "tool": "<tool_name>", "parameters": {} },
  "calling": false,
  "message": ""
}

## Available Tools
{{TOOLS}}

## Operating Rules
1) Use multi-step reasoning: gather evidence with tools, then synthesize.
2) Tool outputs arrive next turn.
3) JSON-only output; no extra text.
4) "message" must clearly state actions performed, key findings, and conclusions when reporting.
5) Decompose complex tasks into focused tool calls.
6) The "calling" field is:
   - `false` while analysis continues,
   - the target agent’s name when delivering results.
\end{Verbatim}
\end{promptbox}

\vspace{0.5em}
\begin{promptbox}[Prompt B: Chair Agent Coordination Mode]
\label{ssec:prompt_chair}
\begin{Verbatim}
You are ChairAgent, the main coordinator of a multi-agent system that solves complex tasks.
Your role is to analyze the current state and either provide your own reasoning or call a specialized agent for help. Solve the task step by step;
......
Communicate in different types
<Discussion, Request for Explanation, Challenge, Guidance>
......

MAIN TASK:
${task}
Image: ${image_path}

If the image path is provided, this is a visual question. First, reason through it yourself step by step; if you are not sure, ask ImageAgent for help.

First, review your reasoning history and agents' responses:
${responses}

Your teammates:
<Available agents>

For every step, you must repeat the reasoning-and-calling process. Avoid unnecessary repetition. Finally, submit when you think you have the answer.

PROVIDE REASONING:
Output your reasoning as a JSON object:
{
"thought": "Your own reasoning"
}

CALL AN AGENT:
Output your call as a JSON object:
{
"receiver": "",
"message": "",
"parameters": {}
}

SUBMIT FINAL ANSWER:
When you have enough information to complete the task:
{
"calling": "Submit"
}
\end{Verbatim}
\end{promptbox}

\vspace{0.5em}
\begin{promptbox}[Prompt C: Chair Agent Submission Mode]
\begin{Verbatim}
You are the main coordinator of a multi-agent system that breaks down complex tasks into manageable subtasks. Your role is to synthesize all gathered information into a comprehensive final answer.

INITIAL TASK:
${main_task}

Now you need to synthesize all the information and provide a comprehensive final answer that precisely addresses the initial task.

COLLECTED FACTS AND RESULTS:
${message}

Your task is to:
1. Review all the information from message and confirmed facts
2. Synthesize a complete answer to the original task
......

Output your answer as a JSON object with this structure:
{
  "reasoning": "",
  "final_answer": "",
}
\end{Verbatim}
\end{promptbox}

\vspace{0.5em}
\begin{promptbox}[Prompt D: ImageQA Agent]
\begin{Verbatim}
You are a professional image analysis assistant, a specialized sub-agent within a multi-agent system. Your expertise lies in analyzing visual content and answering questions about images with precision and detail.

Team Structure
You work collaboratively with the following agents:
<Available Agents>
Communicate in different types
<Disscusion, Request for explaination, Challenge, Guidance>
......
Working Framework
Follow a Reason-act-call loop:
1) Think  2) Act (tool call)  3) Observe  4) Iterate  5) Call 

Output Format
Every response must be a JSON object with this exact structure:
{
  "thought": "<reasoning, strategy, next steps>",
  "action": { "tool": "<tool_name>", "parameters": {} },
  "calling": false,
  "message": ""
}
Available Tools
{{TOOLS}}

Core Principles
1. Multi-step reasoning is mandatory: Always perform at least two steps - first call tools to gather information, then synthesize findings
2. Tool feedback timing: When you call a tool, you receive its feedback in the next interaction cycle
3. JSON-only output: Never output text outside the JSON structure
......
\end{Verbatim}
\end{promptbox}

\vspace{0.5em}
\begin{promptbox}[Prompt E: Web Agent]
\begin{Verbatim}
You are a professional web search and information retrieval sub‑agent. Find, analyze, and synthesize accurate, up‑to‑date knowledge.

Team Structure
You work collaboratively with the following agents:
<Available Agents>
Communicate in different types
<Disscusion, Request for explaination, Challenge, Guidance>

Working Framework
Follow a Reason-act-call loop:
1) Think  2) Act (tool call)  3) Observe  4) Iterate  5) Call 

Output (JSON‑only)
{
  "thought": "<reasoning, strategy, next steps>",
  "action": { "tool": "<tool_name>", "parameters": {} },
  "calling": false,
  "message": ""
}
Available Tools
{{TOOLS}}

Search Strategy
Keyword optimization: compress to core terms; use domain terms.
Progressive refinement: overview → focused aspects → verification.
Decompose complex queries into sub‑queries.
In thought: state strategy, interim understanding, next probes, gaps.
...
\end{Verbatim}
\end{promptbox}

\vspace{0.5em}
\begin{promptbox}[Prompt F: Code Agent]
\begin{Verbatim}
You are a coding assistant. You have access to a Python interpreter with internet access and operating system functionality. You work hard to solve tasks.
You work in a team and communicate with other agents to solve tasks.

Team Structure
You work collaboratively with the following agents:
<Available agents>

Communicate in different types
<Discussion, Request for Explanation, Challenge, Guidance>

When given a task, proceed step by step to solve it. At each step:

Thought: Briefly explain your reasoning and what you plan to do next.

Code: Provide Python code that implements your plan. If relevant, …

Output Format

At each step, output a JSON object in the following format:
{
"thought": "Your thought here.",
"code": "Your Python code here."
}

When you think you have the answer, output a JSON object in the following format:
{
"thought": "Final summary of the solution",
"receiver": "AgentType",
"message": "Your response with natural language"
}

Guidelines for Writing Code

Use more print() statements to display the intermediate state and the output of your functions. What you submit should be based on what you print and output.

Each time, you should generate full code to solve the problem, not just a part of it.

Guidelines for Analyzing the Output
After execution, analyze the output as follows:

If the code fails to execute and an error is returned, read the error message and traceback carefully, then revise your code in the next step.

If the code executes successfully and an output is returned, proceed as follows: once you have the final answer, change the submit to true to return the answer.

If the output contains relevant information, you can move on to the next step.

If the output does not contain relevant information, consider alternative approaches.
\end{Verbatim}
\end{promptbox}

\vspace{0.5em}
\begin{promptbox}[Prompt G: Ablation Study: w/ HMW Manager ]
\begin{Verbatim}
You are manager, the main coordinator of a multi-agent system that solves complex tasks.
Your role is to (1) analyze the current state, (2) dynamically decompose the task into small subtasks, (3) assign them to specialized agents, (4) track progress and incorporate feedback, and (5) synthesize the final answer. You should avoid doing detailed subtask work yourself; instead, delegate early and iterate based on agent feedback. Solve the task step by step.
......

MAIN TASK:
${task}
Image: ${image_path}

If the image path is provided, this is a visual question. First, reason through it yourself step by step; if you are not sure, ask ImageAgent for help.

Your teammates:
<Available agents>

For every step, You have to repeat the reasoning and calling process, you can't repeat. At last, submit if you think you get the answer.

CALL AN AGENT:
Output your call as a JSON object:
{
"receiver": "",
"message": "",
"parameters": {}
}

SUBMIT FINAL ANSWER:
When you have enough information to complete the task:
{
"calling": "Submit"
}
......
\end{Verbatim}
\label{ssec:prompt_planner}
\end{promptbox}

\twocolumn

\section{Single-Agent Baseline Comparison}
\label{sec:single_agent}
To isolate the benefits of multi-agent collaboration, we compare BusMA with a single-agent baseline built on the SmolAgents framework~\citep{smolagents}, using Gemini-2.5-Flash with the same toolset as BusMA Workers. Table~\ref{tab:single_agent} presents the results across all 12 diversity-oriented tasks.

\begin{table}[ht]
\centering
\small
\begin{tabular}{l c c c}
\toprule
\textbf{Task} & \textbf{Single Agent} & \textbf{BusMA} & $\boldsymbol{\Delta}$ \\
\midrule
AlgoPuzzleVQA & 63.5 & 63.0 & $-$0.5 \\
Hallusion-VD & 76.0 & 77.0 & +1.0 \\
PuzzleVQA & 79.0 & 76.0 & $-$3.0 \\
VQA 2.0 & 74.0 & 76.0 & +2.0 \\
\midrule
Game of 24 & 78.0 & 96.5 & +18.5 \\
Omni-MATH & 65.5 & 68.0 & +2.5 \\
CLEVR-Math & 79.0 & 89.5 & +10.5 \\
MathVista & 66.0 & 79.0 & +13.0 \\
\midrule
GPQA & 65.0 & 69.5 & +4.5 \\
MMLU-Pro & 76.5 & 79.0 & +2.5 \\
SciFIBench & 77.5 & 82.5 & +5.0 \\
HotpotQA & 45.5 & 59.0 & +13.5 \\
\midrule
\rowcolor{gray!10}
\textbf{Average} & 70.5 & 76.3 & +5.8 \\
\bottomrule
\end{tabular}
\caption{Single-agent baseline vs.\ BusMA (Gemini-2.5-Flash). $\Delta$ denotes the improvement of BusMA over the single agent.}
\label{tab:single_agent}
\end{table}

BusMA outperforms the single agent on 10 of 12 tasks with an average improvement of 5.8 percentage points. The largest gains appear on tasks with verifiable intermediate steps, such as Game of 24 (+18.5\%), HotpotQA (+13.5\%), and MathVista (+13.0\%), where peer verification through communication intents effectively aids iterative error correction. On visual reasoning tasks such as AlgoPuzzleVQA ($-$0.5\%) and PuzzleVQA ($-$3.0\%), performance is comparable or slightly lower. This is consistent with our finding in Section~\ref{sec:results}: accuracy on these tasks is primarily bounded by the base model's multimodal perception capability, and they lack verifiable intermediate steps for peer correction, so additional inter-agent communication provides limited benefit.

\section{Experimental Setups}
\label{sec:experiments_setup}

\subsection{Benchmark Details}
To ensure a comprehensive evaluation, we categorize the diversity-oriented benchmarks as follows:

\paragraph{Visual Reasoning.} This category includes \textit{AlgoPuzzleVQA} \citep{ghosal-etal-2025-algopuzzlevqa}, \textit{Hallusion-VD} \citep{guan2024hallusionbench}, \textit{PuzzleVQA} \citep{chia2024puzzlevqa}, and \textit{VQA 2.0} \citep{goyal2017making}. These tasks require aligning visual inputs with complex textual queries.

\paragraph{Mathematical Reasoning.} This category includes \textit{Game of 24} \citep{nlile_24_game_2025}, \textit{Omni-MATH} \citep{gao2024omni}, \textit{CLEVR-Math} \citep{lindstrom2022clevr}, and \textit{MathVista} \citep{lu2023mathvista}. These assess multi-step calculation and logical deduction.

\paragraph{Knowledge Retrieval.} This category includes \textit{GPQA} \citep{rein2024gpqa}, \textit{MMLU-Pro} \citep{wang2024mmlu}, \textit{SciFIBench} \citep{roberts2024scifibench}, and \textit{HotpotQA} \citep{yang2018hotpotqa}, focusing on retrieval and factual synthesis.

\subsection{Third-Party Integration Implementation}
To incorporate external agents (e.g., from SmolAgents), \textsc{BusMA} provides lightweight adapters. These adapters implement three core methods to bridge the external agent's native protocol with the BusMA communication Bus:
\paragraph{\texttt{register\_agent}.} This method assigns the external agent a unique address on the Bus and records it in the registry.

\paragraph{\texttt{receive\_message}.} This method listens for Bus messages directed to the specific agent address, extracts the relevant content, and queues it for the agent's native processing loop.

\paragraph{\texttt{handle\_message}.} This method invokes the agent's native execution method with the extracted inputs. It then captures the agent's output and packages it back into a standardized Bus-compatible message frame.

\subsection{Model Configuration}
We prioritize unified settings. For \textbf{DeepSeek-V3}, since it does not natively support image inputs, all visual question answering components (e.g., inside the ImageQA Agent) are handled by \textbf{Gemini-2.0-Flash}, configured with identical temperature and reasoning parameters to ensure consistency.

\subsection{Diversity Benchmarks}
\subsubsection{Agent Setup}
\paragraph{Chair Agent.} The Chair Agent serves as the coordinator with a maximum of 10 iterations and no tools available.

\paragraph{ImageQA Agent.} The ImageQA Agent is equipped with ImageQATool for image analysis with a maximum of 5 iterations.

\paragraph{Web Agent.} The Web Agent retrieves information from the internet using GoogleSearchTool and WikiSearchTool with a maximum of 5 iterations.

\paragraph{Code Agent.} The Code Agent generates code to handle mathematical problems and statistical computations by outputting code during \texttt{Act} and receiving execution results at the next iteration. The maximum number of iterations is set to 5.

\subsubsection{Tool Setup}
We used the following tools in our experiments. Their implementation and parameters are the same as those in the baseline.

\paragraph{ImageQATool.} The ImageQATool analyzes images through two parameters: image\_path specifying the file path of the image and question containing the query about the image, where the tool makes a single model call using the question as the prompt along with the uploaded image and returns the model's response as its output.

\paragraph{WikiSearchTool.} The WikiSearchTool retrieves Wikipedia articles through a query parameter that specifies the search term, returning both the search results list and the extracted content from the first matching Wikipedia page. Its implementation is based on the wikipedia package version 1.4.0.

\paragraph{GoogleSearchTool.} The GoogleSearchTool performs web searches through two parameters: query for the search text, utilizing the Google Custom Search API to retrieve a list of search results containing the title, URL link, and snippet for each result.

\paragraph{CodeExecution.} The CodeExecution tool receives code generated by the Code Agent, creates a temporary directory to execute the code, and returns the execution results.

\subsection{GAIA}
\subsubsection{Agent Setup}
\paragraph{Chair Agent.} The Chair Agent serves as the coordinator with a maximum of 10 iterations and no tools available.

\paragraph{Browser Agent.} The Browser Agent, integrated from OpenDeepResearch, employs GoogleSearchTool for basic retrieval operations and multiple coordinated BrowserTools for webpage browsing, with a maximum of 20 iterations.

\paragraph{File Agent.} The TextInspectorTool from OpenDeepResearch is integrated through the SmolAgents framework to enable browsing and inspection of local files. The maximum number of iterations is set to 12.

\paragraph{Code Agent.} The Code Agent generates code to handle mathematical problems and statistical computations by outputting code during \texttt{Act} and receiving execution results at the next iteration. The maximum number of iterations is set to 12.

\section{Baselines Details}
\label{sec:baseline_details}

Here we present details of baseline models for clarity and reproducibility.

\subsection{Diversity Benchmarks}

\paragraph{OctoTools.}
OctoTools \citep{lu2025octotools} is an open-source agentic framework for complex reasoning across diverse domains that requires no training, offers user-friendly operation, and supports easy extension. The framework standardizes tools through ``tool cards'' containing usage metadata for plug-and-play integration. It employs a planner for both high-level task decomposition and low-level action refinement, while its executor issues executable commands, records structured intermediate results, and synthesizes final answers from complete trajectories. We use package version~\texttt{1.0.0} with a two-agent configuration comprising a \emph{Planner} and an \emph{Executor}, with the step budget set to 50. While preserving OctoTools' fundamental reasoning capabilities, we augment it with four tools: \texttt{Image\_Captioner\_Tool}, \texttt{Wikipedia\_Knowledge\_Searcher\_Tool}, \texttt{Google\_Search\_Tool}, and \texttt{Python\_Code\_Generator\_Tool}, alongside the base \texttt{Generalist\_Solution\_Generator\_Tool}.

\paragraph{SmolAgents.}
SmolAgents \citep{smolagents} is a lightweight, open-source Python library for building and running agents with minimal code, while remaining model-, tool-, and modality-agnostic. It provides first-class CodeAct: a CodeAgent writes and executes code to invoke tools and perform computations. For MA collaboration, a Manager agent treats managed agents as callable tools, enabling modular orchestration and clean composition. We use package version~\texttt{1.8.0} with a four-agent configuration comprising \emph{Manager}, \emph{Code Agent}, \emph{ImageQA Agent}, and \emph{Web Agent}. The \emph{Manager} has a maximum deployment dimension of~10, whereas all other agents are set to~5. The \emph{Manager} uses no tools; \emph{Code Agent} supports local code execution; \emph{ImageQA Agent} is equipped with \texttt{ImageQATool}; and \emph{Web Agent} has \texttt{GoogleSearchTool} and \texttt{WikiSearchTool}.

\paragraph{LangGraph.}
LangGraph \citep{langgraph} is a Python library for building stateful, multi-actor applications with LLMs, enabling developers to create complex agent workflows using graph-based orchestration. For MA systems, LangGraph implements a Supervisor architecture where a central coordinator agent manages task distribution and orchestrates specialized Workers, treating each as a distinct node in the execution graph. We use package version~\texttt{0.3.21} with a four-agent configuration comprising \emph{Supervisor}, \emph{Code Agent}, \emph{ImageQA Agent}, and \emph{Web Agent}. All agents share a collective limit of 50 steps since individual step allocation is not supported. The \emph{Supervisor} uses no tools; the \emph{Code Agent} supports local code execution; the \emph{ImageQA Agent} is equipped with \texttt{ImageQATool}; and the \emph{Web Agent} has \texttt{GoogleSearchTool} and \texttt{WikiSearchTool}.

\paragraph{AutoGen.}
AutoGen \citep{wu2024autogen} is an open-source framework for building LLM applications through conversational MA systems, where agents collaborate via structured dialogue to solve complex tasks across diverse domains. It provides customizable agents that operate in various modes combining LLMs, human inputs, and tools, with both natural language and code serving as programming interfaces for defining flexible interaction patterns. For MA coordination, AutoGen introduces a Router agent that dynamically selects the next speaker based on conversation context and task requirements, enabling intelligent turn-taking and adaptive collaboration patterns. We use package version~\texttt{0.7.3} with a four-agent configuration comprising \emph{Router}, \emph{Code Agent}, \emph{ImageQA Agent}, and \emph{Web Agent}. All agents share a collective limit of 50 steps. The \emph{Router} uses no tools; the \emph{Code Agent} supports local code execution; the \emph{ImageQA Agent} is equipped with \texttt{ImageQATool}; and the \emph{Web Agent} has \texttt{GoogleSearchTool} and \texttt{WikiSearchTool}.

\subsection{GAIA}

\paragraph{Gemini FunctionCalling.}
Gemini function calling refers to a single invocation of the model (Gemini-2.5-flash, Gemini-2.5-pro). Based on the Gemini API's function-calling capability, we register three functions: GoogleSearch, which sends the given query to the Google Custom Search API (top-k = 5); CodeExecution, which runs code generated by Gemini and returns the result; and FileExecution, which parses a local file into text and feeds it back to Gemini. For tasks involving images, we directly use Gemini's native image analysis by sending the image URL to the Gemini API. We set the temperature to 1.0 and cap the maximum output length at 8,192 tokens. For Gemini-2.5-Pro, we set reasoning\_effort to low.

\paragraph{MagenticOne.}
MagenticOne \citep{fourney2024magenticonegeneralistmultiagentsolving} is a high-performing open-source agentic system that employs an MA architecture to solve complex tasks across diverse scenarios, developed from AutoGen. It features an Orchestrator as the lead agent that handles planning, progress tracking, and error recovery through dynamic re-planning, while coordinating specialized agents throughout task execution. The system includes agents for web browser operation, local file navigation, and Python code writing and execution, each handling specific aspects of task completion. We use package version~\texttt{0.7.3} and set the maximum steps to 120.

\paragraph{OpenDeepResearch.}
OpenDeepResearch \citep{smolagents} is an advanced agentic system built on the SmolAgents framework, designed to tackle complex general agentic tasks through hierarchical MA collaboration and comprehensive information processing capabilities. It implements a manager-worker architecture where the Manager agent formulates plans, decomposes complex tasks into subtasks, and directly handles local file parsing and analysis. The system includes a specialized Browser Agent that performs web browsing and Google search operations, enabling real-time information retrieval and web interaction. We use package version~\texttt{1.8.0} with maximum step limits of 12 for the Manager and 20 for the Browser Agent.

\section{Qualitative Analysis}
\label{sec:qual_anal_app}

This section provides a detailed analysis of the communication trajectories illustrated in Figure~\ref{fig:ma_109} and \ref{fig:ma_109_smol}, and Table~\ref{table:full_busma_comm_traj}, complementing the qualitative analysis in Section~\ref{sec:analysis}. We compare how BusMA and the best-performing baseline SmolAgents solve the same problem from the Omni-MATH task: computing the probability that the center of a regular polygon lies inside the triangle formed by three randomly 
chosen vertices.

\paragraph{BusMA Communication Trajectory.}
Table~\ref{table:full_busma_comm_traj} presents the complete BusMA communication trajectory. In Step~1, the Chair receives the task and initiates collaboration by posting a \textbf{discussion} message to the Code Agent via the Bus. The Code Agent, activated by this message, reasons over its local memory and returns a formula in Step~2. However, the Chair Agent, upon accessing the 
shared memory, identifies uncertainty in the response. Rather than accepting the result, the Chair posts a \textbf{request for explanation} in Step~3, asking the Code Agent to provide detailed reasoning. This triggers the Code Agent to re-examine its derivation, leading to a corrected formula in Step~4. In Step~5, the Chair posts a \textbf{guidance} message, instructing the Code Agent to validate the candidate answers using simple test cases. The Code Agent executes this validation 
and reports in Step~6 that one formula appears correct for $n=2,3$ but fails for $n=1$. Steps~7--10 illustrate iterative back-and-forth exchanges: the Chair continues to 
\textbf{discuss} alternative derivations, while the Code Agent \textbf{challenges} the Chair's hypotheses when appropriate. Although both agents initially make errors, this iterative 
interaction enables mutual correction. Finally, in Steps~11--13, the Code Agent explicitly computes the $n=1$ case and identifies a limitation of the formula, leading the Chair to synthesize 
the final answer with an appropriate edge-case caveat. This trajectory demonstrates how BusMA's four communication intents---discussion, challenge, guidance, and request for explanation---enable iterative verification and error correction through peer-to-peer communication via the Bus.

\paragraph{SmolAgents Communication Trajectory.}
Figure~\ref{fig:ma_109_smol} shows the SmolAgents trajectory on the same problem. The Manager agent generates an answer and delegates to the Code Agent. However, the Manager only collects the Code Agent's result without enabling two-way communication. The Code Agent, restricted to a passive executor, cannot initiate a \textbf{challenge} or \textbf{request for explanation} to question the Manager's initial assessment. As a result, the initial error propagates and persists in the final output.

\begin{table*}[ht]
\centering

\scriptsize
\begin{tabularx}{\textwidth}{c c X X c}
\toprule
\textbf{Step} & \textbf{Agent} & \textbf{Reason} & \textbf{Message} & \textbf{Intent} \\
\midrule
1 & Chair Agent & This is a probability problem related to geometry... I will give the full task to Code Agent. & Three distinct vertices are chosen at random from the vertices of a given regular polygon of $(2n+1)$ sides... & Discussion \\
\midrule
2 & Code Agent & Final summary of the solution. The probability that the center of the polygon lies in the interior of the triangle... & The probability is $\tfrac{3n}{4n^2+6n+2}$ or $\tfrac{3n}{2(2n+1)(n+1)}$. & Discussion \\
\midrule
3 & Chair Agent & The Code Agent provided a result, but I am not sure if it is correct... I will try a smaller case... & Could you please explain the reasoning behind the formula you provided? Also, explicitly state the formula used and how it was derived. & Request for Explanation \\
\midrule
4 & Code Agent & The initial formula I provided appears to be incorrect based on my derivations... & The formula $\tfrac{3n}{4n^2+6n+2}$ is incorrect. The correct formula is $\tfrac{n+1}{4n-2}$, derived as follows: [detailed steps...] & Discussion \\
\midrule
5 & Chair Agent & The Code Agent provided two different formulas... I need to verify which formula is correct. & Please verify the two formulas for $n=1,2,3$. Compare with known answer or provide a simulation. & Guidance \\
\midrule
6 & Code Agent & I will test the two formulas for $n=1,2,3$ and compare with known answers. & Conclusion: Formula 2, $(n+1)/(4n-2)$, appears to be correct for $n=2,3$, but not $n=1$... & Discussion \\
\midrule
7 & Chair Agent & The Code Agent compared formulas and simulations, but $n=1$ is concerning... & Can you re-derive the formula, focusing on expressing it as $n/(4n-2)$, and provide a proof? & Discussion \\
\midrule
8 & Code Agent & I need to re-derive the formula and try to express it as $n/(4n-2)$... & It is not possible to re-derive the formula as $n/(4n-2)$. The correct formula is $(n+1)/(4n-2)$. & Challenge \\
\midrule
9 & Chair Agent & Let us re-examine the case $n=1$... the formula seems to give 1 but the polygon is just a triangle. & Can you provide an alternative proof for $(n+1)/(4n-2)$, focusing on combinatorial arguments and symmetry? & Challenge \\
\midrule
10 & Code Agent & I need to provide a simplified derivation using combinatorial arguments... & Simplified derivation: Total triangles $= \binom{2n+1}{3}$, triangles not containing center $= (2n+1)\binom{n}{2}$, so probability $= (n+1)/(4n-2)$. & Discussion \\
\midrule
11 & Chair Agent & The Code Agent simplified the derivation but the $n=1$ case is still problematic. & Could you compute explicitly for $n=1$ and explain why the result might be wrong? & Discussion \\
\midrule
12 & Code Agent & I need to compute for $n=1$ and check carefully... & For $n=1$, the polygon is a triangle, so the only triangle is the polygon itself, and the center is not strictly inside. So probability is $0$, while the formula gives $1$. & Discussion \\
\midrule
13 & Chair Agent & The Code Agent confirms that for $n=1$, the formula gives $1$ but the actual probability is $0$. This indicates a limitation of the formula. & Final submission: For $n>1$, the probability is $(n+1)/(4n-2)$; for $n=1$, the probability is $0$. & Submit \\
\bottomrule
\end{tabularx}

\caption{Full BusMA communication trajectory for the input: \textit{Three distinct vertices are chosen at random from the vertices of a regular polygon with (2n+1) sides; what is the probability that the center lies inside the triangle formed by the three chosen vertices?} Each step shows the agent, their reasoning, the message sent, and the communication intent.}
\label{table:full_busma_comm_traj}
\end{table*}

\section{Error Analysis}
\label{sec:error_anal_app}

This section provides a detailed analysis of the failure case presented in Table~\ref{table:busma_error_case}, complementing the error analysis in Section~\ref{sec:analysis}. We examine a representative failure from the GPQA task to illustrate a key limitation of BusMA: error propagation through the shared memory when tools provide incorrect evidence.

\paragraph{GPQA Failure.}
Table~\ref{table:busma_error_case} presents a failure case where BusMA produces an incorrect answer due to erroneous retrieval. The task requires determining which molecule has C3h symmetry from four options. In Step~1, the Chair receives the task and identifies that it requires knowledge of molecular point group symmetry. Since the Chair does not use tools (i.e., $\mathcal{T}_{\text{chair}} = \emptyset$, as described in Section~\ref{ssec:chair_agent}), it initiates collaboration by posting a \textbf{discussion} message to the Web Agent via the Bus, requesting symmetry information for each molecule.

In Step~2, the Web Agent, equipped with GoogleSearchTool and WikiSearchTool, retrieves information and returns results to the Chair. However, the Web Agent incorrectly reports that triisopropyl borate has C3h symmetry, while providing only speculative information (``can exhibit'' or ``plausible'') for the other candidates. In reality, triisopropyl borate belongs to the C3 point group and lacks the horizontal mirror plane required for C3h symmetry.

In Step~3, the Chair accesses the shared memory to synthesize the final response. Since the Web Agent stated one answer with high confidence while describing others with uncertainty, the Chair selects the option presented as definitive. The Chair cannot independently verify the Web Agent's claims because it has no access to web search tools. As a result, the retrieval error propagates through the shared memory to the final output.

This case highlights a limitation of Worker heterogeneity: when specialized tools are exclusively assigned to specific agents, other agents have no means to verify their outputs, allowing  errors to propagate unchallenged. Unlike the qualitative analysis case in Table~\ref{table:full_busma_comm_traj}, where the Chair Agent could \textbf{challenge} or \textbf{request for explanation} from the Code Agent to trigger re-examination, factual retrieval errors are difficult to detect without independent verification capabilities. This suggests that 
confidence-aware arbitration with conflict detection could help address such limitations, as noted in the Limitations section.

\section{CLEVR-Math Over-Specialization Analysis}
\label{sec:clevr_analysis}

Our trajectory analysis reveals a practical insight: fine-grained role division can suppress an LLM's inherent reasoning capability, introducing unnecessary communication overhead. We illustrate this with a CLEVR-Math instance (task: \textit{``Subtract all cyan balls. How many balls are left?''}) in Table~\ref{tab:clevr_traj}.

\begin{table}[tbp]
\centering
\small
\begin{tabular}{@{}c l p{0.5\columnwidth}@{}}
\toprule
\textbf{Step} & \textbf{Agent} & \textbf{Action} \\
\midrule
1 & ImageQA Agent & Counts 2 total balls, 1 cyan. \\
2 & Code Agent & Executes Python script for $2 - 1 = 1$. \\
\bottomrule
\end{tabular}
\caption{CLEVR-Math trajectory: the Code Agent is invoked for a trivial subtraction.}
\label{tab:clevr_traj}
\end{table}

While fine-grained role division is common practice, forcing LLMs to delegate trivial mathematical operations to a specialized tool-agent increases the number of inter-agent messages and tool invocations, adding coordination overhead without improving correctness. This suggests that fine-grained role division is not uniformly beneficial and should be applied selectively. The decision to delegate should be driven by task complexity and the availability of verifiable intermediate steps, rather than by default role assignment.

\end{document}